\documentclass[runningheads]{llncs}

\usepackage{algorithm}
\usepackage{algpseudocode}
\usepackage{amsmath,amssymb,amsfonts}
\usepackage{array}
\usepackage{multirow}
\usepackage{microtype}
\usepackage{graphicx}
\usepackage{tabularx}
\usepackage[utf8]{inputenc}
\usepackage[english]{babel}

\DeclareMathOperator*{\argmin}{arg\,min}

\begin{document}
    \title{Robust Clustering on High-Dimensional Data with Stochastic Quantization}

    \author{Anton Kozyriev\inst{1} \and Vladimir Norkin\inst{1,2}}
    \authorrunning{Anton Kozyriev and Vladimir Norkin}

    \institute{National Technical University of Ukraine ''Igor Sikorsky Kyiv Polytechnic Institute'', Kyiv, 03056, Ukraine \email{a.kozyriev@kpi.ua} \orcidID{0009-0007-6692-2162} \and
    V.M.Glushkov Institute of Cybernetics, National Academy of Sciences of Ukraine, Kyiv, 03178, Ukraine \email{vladimir.norkin@gmail.com, v.norkin@kpi.ua} \orcidID{0000-0003-3255-0405}}

    \maketitle

    \begin{abstract}

This paper addresses the limitations of conventional vector quantization algorithms, particularly K-Means and its variant K-Means++, and investigates the Stochastic Quantization (SQ) algorithm as a scalable alternative for high-dimensional unsupervised and semi-supervised learning tasks. Traditional clustering algorithms often suffer from inefficient memory utilization during computation, necessitating the loading of all data samples into memory, which becomes impractical for large-scale datasets. While variants such as Mini-Batch K-Means partially mitigate this issue by reducing memory usage, they lack robust theoretical convergence guarantees due to the non-convex nature of clustering problems. In contrast, the Stochastic Quantization algorithm provides strong theoretical convergence guarantees, making it a robust alternative for clustering tasks. We demonstrate the computational efficiency and rapid convergence of the algorithm on an image classification problem with partially labeled data, comparing model accuracy across various ratios of labeled to unlabeled data. To address the challenge of high dimensionality, we employ a Triplet Network to encode images into low-dimensional representations in a latent space, which serve as a basis for comparing the efficiency of both the Stochastic Quantization algorithm and traditional quantization algorithms. Furthermore, we enhance the algorithm's convergence speed by introducing modifications with an adaptive learning rate.

\end{abstract}

    \section{Introduction}

Quantization and clustering are fundamental encoding techniques that provide compact representations of original data \cite{Graf_Luschgy_2000,Jain_2010,Scholkopf_Smola_2002}. These techniques have become essential tools for unsupervised learning, with recent applications spanning diverse domains such as location-allocation problems \cite{Wang_Wei_2020}, document classification \cite{Radomirovic_2023,Widodo_2011}, and data compression \cite{Wan_2019}.

In this context, we consider a random variable $\xi$ with values in Euclidean space $\mathbb{R}^n$ and distribution $P(d\xi)$, representing the original distribution. The encoded discrete distribution is parameterized by a set of atoms $\{y_1, \ldots, y_K\}$ with corresponding probabilities $\{q_1, \ldots, q_K\}$. The optimal quantization problem aims to find the encoded distribution that minimizes the distance to the original distribution. This mathematical structure is analogous to the optimal clustering problem, where the objective is to determine the positions of $K$ cluster centers $\{y_1, \ldots, y_K\}$ such that the sum of distances from each element $\xi$ to the nearest cluster center is minimized.

The K-Means algorithm, proposed by Lloyd \cite{Lloyd_1982}, has been widely employed for solving quantization and clustering problems, with numerous extensions \cite{Jain_2010}. Bottou and Bengio \cite{Bottou_1994} interpreted the K-Means algorithm as an analogue of Newton's method and proposed several stochastic gradient descent algorithms for solving optimal quantization and clustering problems. However, traditional clustering algorithms are constrained by the requirement to load all training data into memory, rendering them non-scalable for large datasets. To address this limitation, Sculley \cite{Sculley_2010} introduced the Mini-Batch K-Means algorithm, which utilizes only a small subset of possible $\xi$ values at each iteration.

The Stochastic Quantization algorithm reframes the clustering problem as a stochastic transportation problem \cite{Kuzmenko_Uryasev_2019,Lakshmanan_Pichler_2023} by minimizing the distance between elements of the original distribution $\{\xi\}$ and atoms of the encoded discrete distribution $\{y_k\}$. This approach employs Stochastic Gradient Descent (SGD) \cite{ermoliev1976stochastic,kiefer1952stochastic,Robbins_Monro_1951} to search for an optimal minimum, leveraging its computational efficiency in large-scale machine learning problems \cite{Bottou_2010}. The use of stochastic approximation allows the algorithm to update parameters with only one element $\xi$ per iteration, ensuring memory efficiency without compromising convergence to the minimum \cite{Newton_Yousefian_Pasupathy_2018}.

This paper explores advanced modifications of the Stochastic Quantization algorithm, incorporating accelerated variants of SGD \cite{nesterov1983method,Poliak_1987,walkington_2023} and adaptive learning rate techniques \cite{Duchi_2011,kingma2017adam,tieleman2012rmsprop} to enhance convergence speed. Norkin et al. \cite{Norkin_Kozyriev_Norkin_2024} provide a comprehensive comparison of various SGD variants, highlighting their respective advantages and limitations, while also offering convergence speed estimates.

Given that the optimal quantization problem is non-smooth and non-convex, specialized methods are required \cite{Gandikota_Kane_Maity_Mazumdar_2022,Tang_2017,Zhao_Lan_Chen_Ngo_2021}. To validate its convergence, we apply the general theory of non-smooth, non-convex stochastic optimization \cite{Ermoliev_Norkin_2003,Ermolev_Norkin_1998,mikhalevich2024}. While traditional clustering algorithms lack theoretical foundations for convergence guarantees and rely primarily on seeding techniques \cite{Arthur_Vassilvitskii_2007}, stochastic optimization theory provides specific conditions for the local convergence of the Stochastic Quantization algorithm, which we supplement in this research.

This paper introduces a novel approach to address semi-supervised learning challenges on high-dimensional data by integrating the Stochastic Quantization algorithm with a deep learning model based on the Triplet Network architecture \cite{Hoffer_2015}. The proposed method encodes images into low-dimensional representations in the $\mathbb{R}^3$ latent space, generating meaningful encoded features for the Stochastic Quantization algorithm. By employing the Triplet Network, this approach overcomes the limitations of quantization and clustering algorithms in high-dimensional spaces, such as visualization difficulties and decreased precision as the number of dimensions increases \cite{Kriegel_Kröger_Zimek_2009}.

To demonstrate the efficiency and scalability of the Stochastic Quantization algorithm, we conducted experiments on a semi-supervised image classification problem using partially labeled data from the MNIST dataset \cite{lecun2010mnist}. The Triplet Network is initially trained on the labeled portion of the dataset as a supervised learning model. Subsequently, the trained network is utilized to project the remaining unlabeled data onto the latent space, which serves as input for training the Stochastic Quantization algorithm. The performance of the proposed solution is evaluated using the F1-score metric \cite{Chinchor_1992} for multi-label classification across various ratios of labeled to unlabeled data.

    \section{Stochastic Quantization}

Stochastic Quantization (SQ) represents a paradigm shift from traditional clustering methods by conceptualizing the feature set $\Xi = \{\xi_i, \;i = 1, \ldots, I\}$ and cluster centers $Y = \{y_k, \;k = 1, \ldots, K\}$ as discrete probability distributions. This approach employs the Wasserstein (or Kantorovich-Rubinstein) distance to minimize distortion between these distributions when representing a continuous distribution by a discrete one \cite{Kuzmenko_Uryasev_2019,Lakshmanan_Pichler_2023}. Recent research has extended the application of quantization algorithms to optimal allocation problems for service centers, where atoms of the discrete distribution represent facility and customer locations \cite{Kuzmenko_Uryasev_2019,Norkin_Onishchenko_2005}.

\begin{definition}
    \label{Stochastic Quantization}
    Optimal quantization, as defined by \cite{Kuzmenko_Uryasev_2019}, minimizes the weighted sum of distances between elements of the feature set $\{\xi_i\} \subset \mathbb{R}^{n}$ and centers $\{y_k\} \subset \mathbb{R}^{n}$:

    \begin{equation}
        \label{sq-objective-fn:eq}
        \min_{y = \{ y_1, \ldots, y_K \} \in Y^K \subset \mathbb{R}^{nK}} \min_{q = \{ q_1, \ldots, q_K \} \in \mathbb{R}^K_{+}} \min_{x = \{ x_{ik} \geq 0 \}} \sum_{i=1}^I \sum_{k=1}^K d(\xi_i, y_k)^r x_{ik}
    \end{equation}

    \noindent subject to constraints:

    \begin{equation}
        \label{sq-objective-constraints:eq}
        \sum_{k=1}^K x_{ik} = p_i, \quad \sum_{k=1}^K q_k = 1, \quad i = 1, \ldots, I
    \end{equation}

    \noindent where $p_i > 0, \sum_{i=1}^I p_i = 1$ are normalized supply volumes, $x_{ik}$ are transportation volumes, $d(\xi_i, y_k)_p = \| \xi_i - y_k \|_p = (\sum_{j=1}^n | \xi_{ij} - y_{kj} |^p)^{\frac{1}{p}}$ is the $l_p$ norm defining the distance between elements in the objective function (\ref{sq-objective-fn:eq}), $Y \subset \mathbb{R}^{n}$ is a common constraint set for variables $\{y_k, k = 1, \ldots, K\}$, and $n, I, K \in \mathbb{N}$.
\end{definition}

In this study, we employ the Euclidean norm ($p = 2$) as the distance metric, defined as $d(\xi_i, y_k)_2 = \sqrt{\sum_{j=1}^n | \xi_{ij} - y_{kj} |^2}$. The choice of distance metric may vary depending on the problem domain. For instance, the cosine similarity function $d(\xi_i, y_j)_{\text{cos}} = \cos(\xi_i, y_j) = \frac{\xi_i \cdot y_j}{\| \xi_i \| \cdot \| y_j \|}$ is utilized in text similarity tasks \cite{Babic_2020,vor_der_bruck_pouly_2019}, while Kolmogorov and Levy metrics are employed for probability and risk theory problems \cite{Kuzmenko_Uryasev_2019}.

Given that in the optimal plan, all mass at point $\xi_i$ is transported to the nearest point $y_k$, problem (\ref{sq-objective-fn:eq})-(\ref{sq-objective-constraints:eq}) can be reduced to the following non-convex, non-smooth global stochastic optimization problem:

\begin{equation}
    \label{global-sq-objective-fn:eq}
    \min_{y = \{ y_1, \ldots, y_K \} \in Y^K \subset \mathbb{R}^{nK}} F(y_1, \ldots, y_k)
\end{equation}

\noindent where

\begin{equation}
    \label{global-sq-fn-expansion:eq}
    F(y) = F(y_1, \ldots, y_k) = \sum_{i=1}^I p_i \min_{1 \leq k \leq K} d(\xi_i, y_k)^r = \mathbb{E}_{i \sim p} \min_{1 \leq k \leq K} d(\xi_i, y_k)^r
\end{equation}

Here, $\mathbb{E}_{i \sim p}$ denotes the expected value over the random index $i$ that takes values $\{1, \ldots, I\}$ with probabilities $\{p_1, \ldots, p_I\}$, respectively.

\begin{lemma}
    \label{Lemma 1}
    In the global optimum $y^{*} = (y_1^{*}, \ldots, y_K^{*})$ of (\ref{sq-objective-constraints:eq}), all $\{y_1^{*}, \ldots, y_K^{*}\}$ belong to the convex hull of elements $\{\xi_1, \ldots, \xi_I\}$ in the feature set.
\end{lemma}

\begin{proof}
    The proof proceeds by contradiction. Assume there exists some $y_{k^{*}}^{*} \notin \text{conv}\{\xi_1, \ldots, \xi_I\}$. Let $\bar{y}_{k^{*}}^{*}$ be the projection of $y_{k^{*}}^{*}$ onto $\text{conv}\{\xi_1, \ldots, \xi_I\}$ and consider points $y_{k^{*}}^{*}(\tau) = (1 - \tau)y_{k^{*}}^{*} + \tau\bar{y}_{k^{*}}^{*}$, $\tau \in [0, 1]$. We observe that $\forall\; \xi_i,\; \tau \in (0, 1]: \|y_{k^{*}}^{*}(\tau) - \xi_i\| < \|y_{k^{*}}^{*} - \xi_i\|$. If $\|y_{k^{*}}^{*} - \xi_{i^{*}}\| = \min_{1 \leq k \leq K} \|y_k^{*} - \xi_{i^{*}}\|$ for some $i^{*}$, then

    \begin{equation}
        \min\{\| y_{k^{*}}^{*}(\tau) - \xi_{i^{*}} \|, \min_{k \neq k^{*}} \| y_k^{*} - \xi_{i^{*}} \|\} < \min_k \| y_k^{*} - \xi_{i^{*}} \|
    \end{equation}

    \noindent Thus, $y^{*} = (y_1^{*}, \ldots, y_K^{*})$ is not a local minimum of the objective function (\ref{global-sq-fn-expansion:eq}). Now, consider the case where $\| y_{k^{*}}^{*} - \xi_i \| > \min_k \| y_k^{*} - \xi_i \|$ for all $i$. By assumption, $\min_k \| y_k^{*} - \xi_{i^{\prime}} \|$ for some $i^{\prime}$. The vector $y^{\prime} = (y_1^{*}, \ldots, y_{k^{*} - 1}^{*}, \xi_{i^{\prime}}, y_{k^{*} + 1}^{*}, \ldots, y_K^{*})$ satisfies $F(y^{\prime}) < F(y^{*})$, contradicting the assumption that $y^{*}$ is a minimum. This completes the proof.
\end{proof}

For a continuous probability distribution $P(d\xi)$, we can interpret the objective function (\ref{global-sq-objective-fn:eq}) as a mathematical expectation in a stochastic optimization problem \cite{ermoliev1976stochastic,Newton_Yousefian_Pasupathy_2018,Norkin_Kozyriev_Norkin_2024}:

\begin{equation}
    \label{smooth-stoch-opt-problem:eq}
    \min_{y = \{ y_1, \ldots, y_K \} \in Y^K \subset \mathbb{R}^{nK}} 
    \left[F(y_1, \ldots, y_K) = \mathbb{E} f(y, \xi) = \int_{\xi \in \Xi} f(y, \xi) P(d \xi)\right]
\end{equation}

\noindent with 

\begin{equation}
    \label{smooth-stoch-fn-expansion:eq}
    f(y, \xi) =  \min_{1 \leq k \leq K} d(\xi, y_k)^r, 
\end{equation}

\noindent where the random variable $\xi$ may have a multimodal continuous distribution. The empirical approximation of $F(y)$ in (\ref{smooth-stoch-opt-problem:eq}) is:

\begin{equation}
    \label{empirical-stoch-fn-expansion:eq}
    F_N(y) = \frac{1}{N} \sum_{i=1}^N \min_{1 \leq k \leq K} d(\xi_i, y_k)^r
\end{equation}

\noindent where $\{\tilde{\xi}_j,\; j = 1, \ldots, N\}$ are independent, identically distributed initial samples of the random variable $\xi$. If $K = 1$, $Y$ is convex, and $r\geq 1$, then problem (\ref{global-sq-objective-fn:eq}) is unimodal and reduces to a convex stochastic optimization problem:

\begin{equation}
    \label{convex-stoch-opt-problem:eq}
    \min_{y \in Y} [ F(y) =  \mathbb{E}_{\tilde{i} \sim p} d(\xi_{\tilde{i}}, y)^r ]
\end{equation}

However, for $K \geq 2$, the function $f(\xi, y) = \min_{1 \leq k \leq K} d(\xi, y_k)^r, y = (y_1, \ldots, y_K)$ is the minimum of convex functions and thus is non-smooth and non-convex. In terms of \cite{mikhalevich2024,Norkin_1986}, $f(\xi, y)$ is a random generalized differentiable function of $y$, and its generalized gradient set $\partial_y \,f\,(\xi,y)$ can be calculated by the chain rule:

\begin{eqnarray}
    \label{sq-objective-fn-gradient:eq}
    \begin{aligned}
        \partial_y \,f\,(\xi, y) &= \text{conv.hull} \left\{ \left(\underbrace{\overbrace{0, \ldots, 0, g_{k}}^k(\xi), 0, \ldots, 0}_K\right), \;\; k \in S(\xi, y), \;\; 0 \in \mathbb{R}^n \right\}, \\
        S(\xi, y) &= \{ k: \| \xi - y_{k} \| = \min_{1 \leq k^{\prime} \leq K} \| \xi - y_{k^{\prime}} \| \}, \\
        g_{k}(\xi) &= r \| \xi - y_{k} \|^{r - 2} (y_{k} - \xi)
    \end{aligned}
\end{eqnarray}

The expected value function (\ref{global-sq-fn-expansion:eq}) is also generalized differentiable, and the set-valued mathematical expectation $\mathbb{E}_{\xi} \partial_y\, f\,(\xi, y)$ is a generalized gradient set of function $F$ \cite{mikhalevich2024,Norkin_1986}. Vectors $G(\xi) = (0, \ldots, 0, g_k, 0, \ldots, 0), \;k \in S(\xi, y), \;0 \in \mathbb{R}^n,$ are stochastic generalized gradients of the function $F(y_1, \ldots, y_K)$. These gradients can be utilized to find the optimal solution $y^{*}=(y_1^{*},\ldots,y_K^{*})$ of problems (\ref{global-sq-objective-fn:eq}), (\ref{smooth-stoch-opt-problem:eq}) using Stochastic Gradient Descent (SGD) \cite{Robbins_Monro_1951,kiefer1952stochastic,ermoliev1976stochastic,Norkin_Kozyriev_Norkin_2024} (written in coordinate-wise form):

\begin{eqnarray}
    \label{sgd-update-rule:eq}
    \begin{aligned}
        & y^{t+1}_k = \Pi_{Y} (y^t_k - \rho_t g_k(\tilde{\xi}^t)), \;\;\; \Pi_{Y} (\cdot) = \argmin_{y \in Y} \| \cdot - y\|, \\
        & y^0_k \in Y, \quad k=1,\ldots,K;\quad t=0,1,\ldots, \label{sgd-update-rule2:eq}
    \end{aligned}
\end{eqnarray}

\noindent where $t$ is the iteration number, $\rho_t > 0$ is a learning rate (step) parameter, and $\Pi_{Y}$ is the projection operator onto the set $Y$, $\{\tilde{\xi}^t\}$ are independent identically distributed elements of $\Xi$. The iterative process (\ref{sq-objective-fn-gradient:eq})-(\ref{sgd-update-rule:eq}) for finding the optimal element is summarized in Algorithm \ref{sq:alg}.

\begin{algorithm}
    \caption{Stochastic Quantization}\label{sq:alg}
    \begin{algorithmic}[1]
        \Require $\{ \xi_i, \quad i = 1, \ldots, I \}, \rho, Y, K, T, r$
        \State $y^0 = \{ y_k^0, \quad k = 1, \ldots, K \}$ \Comment{Initialize centers}
        \For{$t \in [0, T - 1]$}
            \State $\tilde{\xi} \in \{ \xi_i, \quad i = 1, \ldots, I \}$ \Comment{Sample an element from the feature set}
            \State $y_k^t, k \in S(\tilde{\xi}, y^t)$ \Comment{Find the nearest center}
            \State $g_k^t = r \| \tilde{\xi} - y_k^t \|^{r - 2} (y_k^t - \tilde{\xi})$ \Comment{Calculate a partial gradient}
            \State $y_k^t := \Pi_Y (y_k^t - \rho g_k^t)$ \Comment{Update the nearest center}
        \EndFor
    \end{algorithmic}
\end{algorithm}

The local convergence conditions of the stochastic generalized gradient method for solving problem (\ref{global-sq-objective-fn:eq}) are described in Theorem \ref{Theorem 1}, with the proof provided in \cite{Ermoliev_Norkin_2003,Ermolev_Norkin_1998}.

\begin{theorem}
    \label{Theorem 1}
    \cite{Ermoliev_Norkin_2003,Ermolev_Norkin_1998}. Consider the iterative sequence $\{ y^t = (y_1^t, \ldots, y_K^t) \}$ formed according to (\ref{sgd-update-rule:eq}), (\ref{sgd-update-rule2:eq}). Let $\{ \tilde{\xi}^t,\;t=0,1,\ldots\}$ be independent sample points from the set $\{ \xi_i, i = 1, \ldots, I \}$ taken with probabilities $\{ p_i, i = 1, \ldots, I \}$; step parameters $\{\rho_t\}$ satisfy conditions:

    \begin{equation}
        \label{sq-convergence-cond:eq}
        \rho_t > 0, \quad \sum_{t=0}^{\infty} \rho_t = \infty, \quad \sum_{t=0}^{\infty} \rho_t^2 < \infty.
    \end{equation}

    Denote $F(Y^{*})$ the set of values of $F$ on critical (stationary) points $Y^{*}$ of problem (\ref{global-sq-objective-fn:eq}), where $Y^{*} = \{ y = (y_1, \ldots, y_K): \partial F(y) \in N_Y (y_1) \times \ldots \times N_Y (y_K) \}$ and $N_Y (y_k)$ represents the normal cone to the set $Y$ at point $y_k$. If $F(Y^{*})$ does not contain intervals and the set $Y$ is a convex compact, then with probability one $\{ y^t \}$ converges to a connected component of $Y^{*}$, and the sequence $\{ F(y^t) \}$ has a limit.
\end{theorem}

While SGD is an efficient local optimization algorithm, the ultimate task is to find global minima of (\ref{global-sq-objective-fn:eq}). A straightforward approach to achieve this is to start the algorithm from different initial points. The research in \cite{Norkin_Pflug_Ruszczynski_1998} proposes a stochastic branch and bound method applicable to the optimization algorithm (\ref{sgd-update-rule:eq}). The idea is to sequentially partition the initial problem into regions (with constraint set $Y_1 \times \ldots \times Y_K$) and use upper and lower bounds to refine partitions with the so-called interchanges relaxation to obtain lower bounds:

\begin{eqnarray}
    \label{sq-branch-bound:eq}
    \begin{aligned}
        \min_{\{ y_k \in Y_k \}} F(y_1, \ldots, y_K)
        &\geq& \sum_{i=1}^I p_i \min_{y \in Y_1\times\ldots\times Y_K} \min_{1 \leq k \leq K} d(\xi_i, y_k)^r \\
        &\geq& \sum_{i=1}^I p_i \min_{1 \leq k \leq K} d(\xi_i, \pi_k(\xi_i))^r,
    \end{aligned}
\end{eqnarray}

\noindent where $\pi_k(\xi_i)=\Pi_{Y_k}(\xi_i)$.

\subsection{Adaptive Stochastic Quantization} \label{adap-stoch-quant:sec}

The minimization of the objective function (\ref{global-sq-objective-fn:eq}) presents a non-smooth, non-convex, and large-scale stochastic optimization problem with multiple local extrema. Although the parameter update process based on Stochastic Gradient Descent (SGD) (\ref{sgd-update-rule:eq}) converges under certain conditions (\ref{sq-convergence-cond:eq}), Qian et al. \cite{qian2020} demonstrated that the variance of gradient oscillations increases proportionally with the size of the training sample:

\begin{equation}
    \label{sgd-oscillations:eq}
    \mathbb{V} (g_{\mathcal{B}_k}) \propto \frac{I^2}{b} \mathbb{V} (g_k),
\end{equation}

\noindent where $\mathbb{V}$ represents the variance over a set, $g_{\mathcal{B}_k} = \frac{1}{b} \sum_{i=1}^{b} g_i (\xi_{\mathcal{B}_i})$ is the averaged gradient over a subset $\xi_{\mathcal{B}_i} \subset \Xi$, and $b = | \xi_{\mathcal{B}_i} |$ is the batch size. These oscillations negatively affect algorithmic stability and reduce convergence speed. Although techniques such as manually adjusting the learning rate $\rho > 0$, employing annealing schedules \cite{Robbins_Monro_1951}, or averaging the gradient over a subset can enhance stability, slow convergence remains a critical limitation of SGD, particularly in high-dimensional models \cite{Norkin_Kozyriev_Norkin_2024}.

Polyak \cite{Poliak_1987} proposed Momentum Gradient Descent (also referred to as the ''Momentum'' or ''Heavy Ball Method'') as a modification to SGD, introducing an acceleration multiplier $0 < \gamma < 1$ to the recurrent sequence (\ref{sgd-update-rule:eq}), drawing on the analogy of physical motion under friction:

\begin{equation}
    \label{momentum-update-rule:eq}
    y^{\,t+1} = y^{\,t} + \gamma (y^{\,t} - y^{\,t-1}) - \rho_t \,g^{\,t}, \quad g^{\,t}\in \partial_y F(y^{\,t}),   \quad t = 1,2,\ldots.
\end{equation}

Nesterov \cite{nesterov1983method,walkington_2023} extended this approach by introducing an extrapolation step for more accurate parameter estimation, known as Nesterov Accelerated Gradient (NAG):

\begin{eqnarray}
    \label{nag-update-rule:eq}
    \begin{aligned}
        \tilde{y}^{\,t} &= y^{\,t} - \rho_t g^{\,t}, \quad g^{\,t} \in \partial_y F(y^{\,t}), \\
        y^{\,t+1} &= \tilde{y}^{\,t} + \gamma (\tilde{y}^{\,t} - \tilde{y}^{\,t-1}), \quad t=1,2,\ldots.
    \end{aligned}
\end{eqnarray}

Both methods (\ref{momentum-update-rule:eq}) and (\ref{nag-update-rule:eq}) have been adapted for non-convex, non-smooth optimization problems \cite{mikhalevich2024}. However, despite their improvement in convergence speed, these modifications often face the vanishing gradient problem when applied to sparse data \cite{Bottou_Curtis_Nocedal_2018}. This issue arises due to the fixed learning rate, which applies equal updates to both significant and insignificant model parameters. To address this, Duchi et al. \cite{Duchi_2011} proposed an adaptive learning rate $\tilde{\rho}_k^{t} = \rho_t / \sqrt{G_k^{\,t} + \varepsilon}$, where the learning rate is normalized over the accumulated gradient to prioritize more significant parameters. This approach, known as AdaGrad, is represented by the update rule:

\begin{equation}
    \label{adagrad-update-rule:eq}
    y^{\,t+1}_k = y^{\,t}_k - \frac{\rho_t}{\sqrt{G_k^{\,t} + \varepsilon}} g^{\,t}_k, \quad k=1,\ldots,K,
\end{equation}

\noindent where $G_k = G_{k-1} + g_{k^*}^2$ is the accumulated sum of squared gradients from previous iterations, and $\varepsilon \ll 10^{-8}$ serves as a smoothing term. Although this method mitigates the convergence issue in sparse datasets, it introduces the challenge of the learning rate decaying too quickly, i.e., $\lim_{k \to \infty} | \tilde{\rho}_k | = 0$. To address this limitation, Tieleman et al. \cite{tieleman2012rmsprop} proposed RMSProp, which normalizes the accumulated gradient using a moving average $G_k = \beta G_{k-1} + (1 - \beta) g_{k^*}^2$. This approach uses a stochastic approximation of the expected value $\mathbb{E} G_k$, controlled by an averaging multiplier $0 < \beta < 1$, to prevent the rapid vanishing of the learning rate.

Further advancements were made by Kingma et al. \cite{kingma2017adam}, who developed the ADAM algorithm, incorporating adaptive moment estimation. The first and second moments of the gradient are estimated as follows:

\begin{eqnarray}
    \label{adam-update-rule:eq}
    \begin{aligned}
        m_k^{\,t} &= \beta_1 m_k^{\,t-1} + (1 - \beta_1) g_k^{\,t}, \\
        v_k^{\,t} &= \beta_2 v_k^{\,t-1} + (1 - \beta_2) \|g_k^{\,t}\|^2, \\
        y_k^{\,t+1} &= y_k^{\,t} - \frac{\rho_t}{\sqrt{v_k^{\,t} + \varepsilon}} m_k^{\,t}, \quad k=1,\ldots,K,
    \end{aligned}
\end{eqnarray}

\noindent where $m_k^{\,t}$ represents the adaptive estimate of the first moment (the expected gradient value), and $v_k^{\,t}$ represents the adaptive estimate of the second moment (the variance). The constants $0 < \beta_1 < 1$ and $0 < \beta_2 < 1$ are used as averaging multipliers. Although $m_k^{\,t}$ and $v_k^{\,t}$ may initially be biased, Kingma et al. introduced bias-corrected estimates:

\begin{equation}
    \label{adam-corrected-estimations:eq}
    \bar{m}_k^{\,t} = \frac{m_k^{\,t}}{1 - \beta_1}, \quad \bar{v}_k^{\,t} = \frac{v_k^{\,t}}{1 - \beta_2}, \quad k=1,\ldots,K.
\end{equation}

Norkin et al. \cite{Norkin_Kozyriev_Norkin_2024} provide a comprehensive review of these adaptive optimization techniques, comparing their convergence properties across a range of problem settings.
    \section{Traditional Clustering Model}

The primary objective of the optimal clustering problem is to identify $K$ cluster centers, denoted as $\{ y_1, \ldots, y_K \}$, that minimize the total distance from each data point $\xi$ to its nearest cluster center. Lloyd's K-Means algorithm \cite{Lloyd_1982} is a well-established approach for addressing quantization and clustering challenges, with various extensions reported in the literature \cite{Jain_2010}. Bottou and Bengio \cite{Bottou_1994} conceptualized the K-Means algorithm as an analogue to Newton's method and introduced several stochastic gradient descent (SGD) algorithms aimed at achieving optimal clustering. These stochastic variants of the K-Means algorithm execute iterations based on either a subset of the dataset or a single data point at a time.

\begin{definition}
    \label{K-Means} 
    \cite{Lloyd_1982}. The iterative K-Means algorithm begins with a set of initial cluster centers $\{ y_k^{\,t}, \> k = 1, \ldots, K \}$. The dataset $\{ \xi_i, \> i = 1, \ldots, I \}$ is partitioned into $K$ mutually exclusive groups $\{ I_1^{\,t}, \ldots, I_K^{\,t} \}$, where a point $\xi_i$ is assigned to group $I_k^{\,t}$ if it satisfies:

    \begin{equation}
        \label{kmeans-group:eq}
        \| \xi_i - y_k^{\,t} \|^2 = \min_{1 \leq k^{\,\prime} \leq K} \| \xi_i - y_{k^{\,\prime}}^{\,t} \|^2.
    \end{equation}

    \noindent Let $N_k^{\,t}$ represent the number of points in group $I_k^{\,t}$, with $N_k^{\,t} = 0$ if $I_k^{\,t} = \emptyset$, and $\sum_{k=1}^K N_k^{\,t} = I$. Notably, both $I_k^{\,t}$ and $N_k^{\,t}$ are contingent upon $y^{\,t}$. The K-Means algorithm iteratively updates the cluster centers $y^{\,t+1}$ using the following estimates:

    \begin{equation}
        \label{kmeans-center-estimation:eq}
        y_{k}^{\,t + 1} = \frac{1}{N_k^{\,t}} \sum_{i \in I_k^{\,t}} \xi_i, \quad k = 1, \ldots, K; \quad t = 0, 1, \ldots.
    \end{equation}

    \noindent These vectors can alternatively be expressed as:

    \begin{eqnarray}
        \label{kmeans-center-alt:eq}
        y_{k}^{\,t + 1} &=& y_k^{\,t} - \frac{1}{N_k^{\,t}} \sum_{i \in I_k^{\,t}} (y_k^{\,t} - \xi_i) = y_k^{\,t} - \frac{1}{N_k^{\,t}} \sum_{i \in I_k^{\,t}} y_k^{\,t} + \frac{1}{N_k^{\,t}} \sum_{i \in I_k^{\,t}} \xi_i \nonumber \\
        &=& \frac{1}{N_k^{\,t}} \sum_{i \in I_k^{\,t}} \xi_i, \quad k = 1, \ldots, K; \quad t = 0, 1, \ldots.
    \end{eqnarray}
\end{definition}

In \cite{Bottou_1994}, the expression for the K-Means algorithm (\ref{kmeans-center-alt:eq}) was linked to the Newtonian step for solving a smooth quadratic problem at each iteration:

\begin{equation}
    \label{kmeans-newton-form:eq}
    \min_{y_1,\ldots,y_K}\left[F^t(y) = \frac{1}{2} \sum_{k=1}^K \sum_{i \in I_k^t} || \xi_i - y_k ||^2\right].
\end{equation}

\noindent with a block diagonal Hessian with diagonal elements $1/N_k^{\,t}$ in block $k$. Furthermore, it is clear that (\ref{kmeans-center-alt:eq}) provides the exact analytical solution to the unconstrained quadratic minimization problem (\ref{kmeans-newton-form:eq}) under a fixed partition $\{I_1^{\,t},\ldots,I_K^{\,t}\}$ of the index set $\{1,\ldots,I\}$.

The paper additionally explores stochastic batch and online versions of stochastic gradient methods with a learning rate of $\frac{1}{t + 1}$ to solve a sequence of problems (\ref{kmeans-newton-form:eq}), though without a rigorous convergence analysis.

The initial positions of cluster centers $\{ y_k^0, \> k = 1, \ldots, K \}$ are determined either randomly from $\{ \xi_i, \> i = 1, \ldots, I \}$ or via the K-Means++ algorithm \cite{Arthur_Vassilvitskii_2007,Nguyen_Duong_2018}. The convergence rate to a local optimum using K-Means++ is estimated as $\mathbb{E} [F] \leq 8(\ln k + 2 ) F^{*}$, where $F^{*}$ denotes an optimal solution \cite{Arthur_Vassilvitskii_2007}. Assuming $\{ y_1^0, \ldots, y_k^0 \} \> (k<K)$ initial centers are chosen, the subsequent center $y_{k+1}^0 \> (k+1<K)$ is selected from $\{ \xi_i, \> i = 1, \ldots, I \}$ with probabilities:

\begin{equation}
    \label{kmeans-plus-plus-init:eq}
    q_j = \frac{\min_{1 \leq s \leq k} || \xi_j - y_s^0 ||^2}{\sum_{i=1}^I \min_{1 \leq s \leq k} || \xi_i - y_s^0 ||^2}
\end{equation}

Subsequent cluster center positions $\{ y_1^{\,t}, \ldots, y_K^{\,t} \}$ for $t > 0$ are calculated using Lloyd's original algorithm (\ref{kmeans-group:eq})-(\ref{kmeans-center-estimation:eq}). An alternative discrete optimization-based initialization strategy for $y^{0}$ is suggested in \cite[Sec. 4, Stage 1]{Kuzmenko_Uryasev_2019}.

Sculley \cite{Sculley_2010} highlighted the inefficiency of Lloyd's algorithm's update rule (\ref{kmeans-center-estimation:eq}) for large datasets, due to the time complexity $O(K \cdot I \cdot d)$ associated with the feature set $\{ \xi_i \}$, where $d$ represents the dimensionality of each sample $\xi_i$. To address this, the author proposed the Mini-Batch K-Means modification, which leverages a small random subset $\Xi^{t}\subset \Xi$ at iteration $t$ to perform the algorithm step (\ref{kmeans-center-estimation:eq}).

Consider the generalized clustering problem (\ref{global-sq-objective-fn:eq})-(\ref{global-sq-fn-expansion:eq}) for any $r\geq 1$:

\begin{eqnarray}
    \label{lloyd-update-step:eq}
    \begin{aligned}
        F(y) = F(y_1, \ldots, y_K) &= \sum_{i=1}^I p_i \min_{1 \leq k \leq K} \| y_k - \xi_i \|^r \\
        &= \mathbb{E}_{i \sim p} \min_{1 \leq k \leq K} \| y_k - \xi_i \|^r \rightarrow \min_y. 
    \end{aligned}
\end{eqnarray}

With the objective function (\ref{lloyd-update-step:eq}) being a generalized differentiable function \cite{Norkin_1986}, its optima can be found via a generalized gradient set, derived using the chain rule:

\begin{equation}
    \label{lloyd-grad-set:eq}
    \partial F(y) = \sum_{i=1}^I p_i \; \partial \min_{1 \leq k \leq K} \| y_k - \xi_i \|^r. 
\end{equation}

The dataset $\{ \xi_i, i = 1, \ldots, I \}$ is divided into non-overlapping subsets $I_k$, $k = 1, \ldots, K$, such that $i \in I_k$ if $\| y_k - \xi_i \| = \min_{k^\prime \in \{ 1, \ldots, K \}} \| y_{k^\prime} - \xi_i \|$. Some subsets $I_k$ may be empty, as in scenarios where:

\begin{equation}
    \label{lloyd-empty-set-cond:eq}
    \max_{1 \leq i \leq I} \min_{1 \leq k \leq K} \|y_k - \xi_i\| < \min_{1 \leq i \leq I} \|y_k - \xi_i\|
\end{equation}

In these modifications, the generalized gradient of $F(y)$ is $g(y) = (g_1(y), \ldots, g_K(y))$, where:

\begin{equation}
    \label{lloyd-gen-grad-component:eq}
    g_k(y) = \begin{cases}
        \sum_{i \in I_k} r p_i \|y_{k} - \xi_i\|^{r\,-2}(y_{k} - \xi_i), & I_k \neq \emptyset, \\
        0, & I_k = \emptyset. 
    \end{cases}
\end{equation}

The standard (unconstrained) generalized gradient method for solving problem (\ref{lloyd-update-step:eq}) is expressed as (for $p_i = 1/I$):

\begin{eqnarray}
    \label{lloyd-gen-grad:eq}
    \begin{aligned}
        y_k^{\,t+1} &= y_k^{\,t} - \rho_t \,g_k(y^{\,t}) \\
                    &= \begin{cases}
                        y_k^{\,t} - \rho_t \frac{r}{I} \sum_{i \in I_k^{\,t}} \|y_k^{\,t} - \xi_i\|^{r\,-2}(y_k^{\,t} - \xi_i), & I_k^{\,t} \neq \emptyset, \\
                        y_k^{\,t}, & I_k^{\,t} = \emptyset,
                    \end{cases}
    \end{aligned}
\end{eqnarray}

\noindent where $I_k^{\,t} = \{i: \| y_k^{\,t} - \xi_i \| = \min_{1 \leq  k^{\,\prime} \leq K} \| y_{k^{\,\prime}}^{\,t}-\xi_i \| \} \quad k = 1 , \ldots , K \quad t = 0, 1 , \ldots$. Analogous to the convergence conditions in \ref{sq-convergence-cond:eq}, the generalized gradient method is guaranteed to converge under the conditions:

\begin{eqnarray}
    \label{lloyd-gen-grad-convergence:eq}
    \lim_{t\rightarrow\infty}\rho_t=0,\quad\sum_{t=0}^\infty\rho_t=+\infty.
\end{eqnarray}

Let $N_k^{\,t}$ denote the number of elements in $I_k^{\,t} \neq \emptyset$ at iteration $t$. If $\rho_t$ is chosen to be dependent on $k$, specifically, $\rho_{t\,k} = 0.5 \frac{I}{N_k^{\,t}}$, then for $r=2$, the process (\ref{lloyd-gen-grad:eq}) becomes equivalent to the K-Means update rule (\ref{kmeans-center-estimation:eq}). However, this choice does not ensure method convergence. A more generalized choice could be $\rho_{t\,k} = \rho_t \frac{I}{N_k^{\,t}}$ with $\rho_t \geq 0$ satisfying conditions (\ref{lloyd-gen-grad-convergence:eq}), ensuring $\lim_{t \to \infty} \max_k \rho_{t\,k} = 0$.

\subsection{Stochastic K-Means Algorithm}

Let $\tilde{\xi}^t$ denote a point sampled from the set $\{\xi_i,\;i=1,2,\ldots,I\}$, where $t$ represents the iteration number. We define a stochastic generalized gradient $\tilde{g}(y) = (g_k(y), k = 1, ..., K)$ as follows:

\begin{equation}
    \label{stoch-lloyd-grad-component:eq}
    \tilde{g}_k(y) = \begin{cases}
        r\|y_{k} - \tilde{\xi}^t\|^{r-2} (y_{k} - \tilde{\xi}^t), & k \in S(\tilde{\xi}^t,y),\\
        0, & k \notin S(\tilde{\xi}^t,y)
    \end{cases}
\end{equation}

\noindent where $S(\tilde{\xi}^t,y) = \argmin_{1 \leq k \leq K} \| y_k - \tilde{\xi}^t \|$. The stochastic generalized gradient method for solving the problem (\ref{lloyd-update-step:eq}) is formulated as:

\begin{equation}
    \label{stoch-lloyd-gen-grad:eq}
    y_k^{t+1} = y_k^t - \rho_t \tilde{g}_k(y^t) = \begin{cases}
        y_k^t - \rho_t r\|y_k^t - \xi_i\|^{r-2}(y_k^t - \xi_i), & k \in S(\tilde{\xi}^t,y^t), \\
        y_k^t, & k \notin S(\tilde{\xi}^t,y^t)
    \end{cases}
\end{equation}

\noindent for $k=1,\ldots,K$ and $t=0,1,\ldots$, where the step multipliers $\{\rho_t\}$ satisfy the conditions specified (\ref{sq-convergence-cond:eq}). Recent studies by \cite{Tang_2017} and \cite{Zhao_Lan_Chen_Ngo_2021} have investigated stochastic K-Means algorithms as methods for addressing corresponding non-convex, non-smooth stochastic optimization problems. However, the convergence properties of these algorithms remain an area requiring further rigorous analysis and consideration.

\subsection{Modifications of K-Means algorithm}

The stochastic K-Means algorithm can be alternatively constructed by dividing the iteration process into epochs of length $I$ and employing a shuffling mechanism at each epoch to select random elements $\tilde{\xi}^t$ \cite{bottou2009curiously,montavon2012neural}. 

The robust clustering model involves solving the problem described by equations (\ref{global-sq-objective-fn:eq}) and (\ref{global-sq-fn-expansion:eq}) with parameter $r \in [1,2)$. This choice of $r$ enhances the model's resilience to outliers in both quantization and clustering. Consequently, the stochastic generalized gradients of the objective function must be calculated using equation (\ref{sq-objective-fn-gradient:eq}), resulting in the stochastic clustering algorithm taking the form of equation (\ref{stoch-lloyd-gen-grad:eq}).

An additional consideration is the application of Cesaro trajectory averaging \cite{Bottou_Curtis_Nocedal_2018,montavon2012neural} to complement the sequences in equations (\ref{lloyd-gen-grad:eq}) and (\ref{stoch-lloyd-gen-grad:eq}):

\begin{equation}
    \label{kmeans-trajectory-avg:eq}
    \begin{aligned}
        \bar{y}_k^{\,t+1} &= (1 - \sigma_{t+1}) \bar{y}_k^{\,t} + \sigma_{t+1} y_k^{\,t+1}, \\
        \sigma_{t+1} &= \frac{\rho_{t+1}}{\sum_{s=1}^{t+1} \rho_{s}}, \quad k = 1, \ldots, K, \quad t=0,1,\ldots.
    \end{aligned}
\end{equation}

Convergence conditions for this averaged sequence, studied in \cite{mikhalevich2024}, permit a learning rate $\rho_t$ proportional to $1 / \sqrt{t+1}$. A similar approach for K-Means generated sequences (\ref{kmeans-center-estimation:eq}) aims to average the sequence by the feature set size $N_k$:

\begin{equation}
    \label{kmeans-trajectory-avg-alt:eq}
    \begin{aligned}
        \tilde{y}_k^{\,t+1} &= (1 - \tilde{\sigma}_{t+1}) \tilde{y}_k^{\,t} + \tilde{\sigma}_{t+1} y_k^{\,t+1}, \\
        \tilde{\sigma}_{k, \,t+1} &= \frac{1}{N_k^{\,t+1}} \left/ \sum_{s=1}^{t+1} \frac{1}{N_k^{\,s}}, \quad k = 1, \ldots, K. \right.
    \end{aligned}
\end{equation}

The standard K-Means algorithm necessitates finding the nearest cluster center $\argmin_{1 \leq k \leq K} \| y_k - \xi_i \|$ for each point $\{ \xi_i, \> i = 1, \ldots, I \}$. This operation can be computationally expensive for large $I$. Furthermore, if points $\xi_i$ are sequentially sampled from a continuous distribution, the sample $\{ \xi_i, \> i = 1, 2, \ldots \}$ can potentially be arbitrarily large. The stochastic algorithm (\ref{stoch-lloyd-gen-grad:eq}) mitigates this issue by using only one sampled point ${\xi}^t$ per iteration, thus solving only one problem $\argmin_{1 \leq k \leq K} \| y_k - \tilde{\xi}^t \|$ at iteration $t$. However, following \cite{Sculley_2010}, one can employ a Mini-Batch of $m$ points $\{ {\xi}^t_{i_1^t}, \ldots, {\xi}^t_{i_m^t} \}$ instead of the entire set $\Xi= \{ \xi_i,\;i=1,\ldots,I \}$, where $1 \leq m < I$, to balance computational efficiency and algorithmic performance.

    \section{Data Encoding with Triplet Network}

Our proposed semi-supervised classification approach addresses a critical challenge in clustering high-dimensional data by integrating Contrastive Learning techniques \cite{Hoffer_2015,Khosla_2020} with Stochastic Quantization. While the Stochastic Quantization algorithm (\ref{sgd-update-rule:eq}) effectively mitigates scalability issues for large datasets, it remains susceptible to the ''curse of dimensionality'', a phenomenon common to clustering methods that rely on distance minimization (e.g., K-Means and K-Means++). Kriegel et al. \cite{Kriegel_Kröger_Zimek_2009} elucidated this phenomenon, demonstrating that the concept of distance loses its discriminative power in high-dimensional spaces. Specifically, the distinction between the nearest and farthest points becomes increasingly negligible:

\begin{equation}
    \label{dimensions-precision-ratio:eq}
    \lim_{n \to \infty} \frac{\max d(\xi_i, y_k) - \min d(\xi_i, y_k)}{\min d(\xi_i, y_k)} = 0.
\end{equation}

Our study focuses on high-dimensional data in the form of partially labeled handwritten digit images \cite{lecun2010mnist}. However, it is important to note that this approach is not limited to image data and can be applied to other high-dimensional data types, such as text documents \cite{Radomirovic_2023,Widodo_2011}. While efficient dimensionality reduction algorithms like Principal Component Analysis (PCA) \cite{Abdi_Williams_2010,Deisenroth_Faisal_Ong_2020} exist, they are primarily applicable to mapping data between two continuous spaces. In contrast, our objective necessitates an algorithm that learns similarity features from discrete datasets and projects them onto a metric space where similar elements are grouped into clusters, a process known as similarity learning.

Recent research \cite{MURASAKI_ANDO_SHIMAMURA_2022,Turpault_Serizel_Vincent_2019} has employed a Triplet Network architecture to learn features from high-dimensional discrete image data and encode them into low-dimensional representations in the latent space. The authors proposed a semi-supervised learning approach where the Triplet Network is trained on a labeled subset of data to encode them into latent representations in $\mathbb{R}^n$, and subsequently used to project the remaining unlabeled fraction onto the same latent space. This approach significantly reduces the time and labor required for data annotation without compromising accuracy.

The Triplet Network, introduced by \cite{Hoffer_2015}, is a modification of the Contrastive Learning framework \cite{Khosla_2020}. Its core idea is to train the model using triplets of samples:

\begin{enumerate}
    \item An anchor sample $\xi_i$: a randomly sampled element from the feature set $\Xi$
    \item A positive sample $\xi^{+}_i$: an element with a label similar to the anchor $\xi_i$
    \item A negative sample $\xi^{-}_i$: an element with a label different from the anchor $\xi_i$
\end{enumerate}

\begin{figure}
    \centering
    \includegraphics[width=\textwidth]{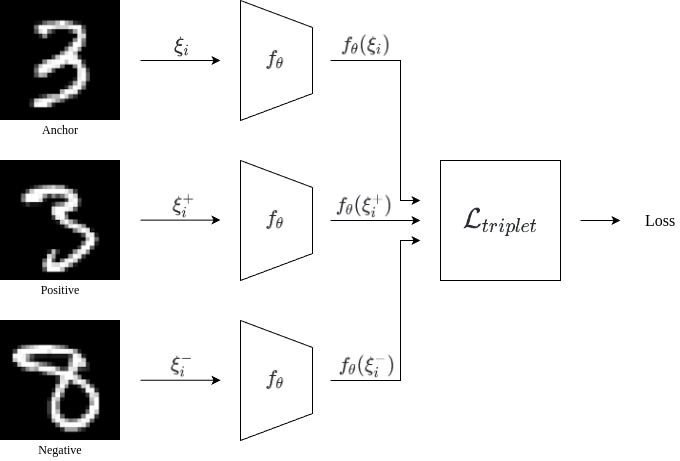}
    \caption{Triplet Network structure for MNIST dataset \cite{lecun2010mnist}.} \label{triplet-network:fig}
\end{figure}

Unlike traditional Contrastive Learning, which compares only positive $\xi^{+}_i$ and negative $\xi^{-}_i$ samples, the Triplet Network learns to minimize the distance between the anchor and positive samples while maximizing the distance between the anchor and negative samples. This is achieved using the triplet loss objective function (see Fig.~\ref{triplet-network:fig}):

\begin{equation}
    \label{triplet-loss-func:eq}
    \mathcal{L}_{triplet}(\theta) = \max (0, d(f_{\theta}(\xi_i), f_{\theta}(\xi^{+}_i)) - d(f_{\theta}(\xi_i), f_{\theta}(\xi^{-}_i)) + \alpha)
\end{equation}

\noindent where $f_{\theta}: \Xi \to \mathbb{R}^n$ is a parameterized abstract operator mapping discrete elements $\Xi$ into latent representations (in our case, a Triplet Network with weights $\theta$), $d: [\mathbb{R}^n, \mathbb{R}^n] \to \mathbb{R}$ is a distance metric between samples, and $\alpha$ is a margin hyperparameter enforcing a minimum separation between positive and negative pairs. Analogous to the Stochastic Quantization distance metric (\ref{sq-objective-fn:eq}), we employed the Euclidean norm $l_2$ for $d(\xi_i, \xi_j)$ in (\ref{triplet-loss-func:eq}).

In our research, we utilized a Convolutional Network architecture as $f_{\theta}$, as proposed by \cite{Lecun_1998}. The detailed overview of the architecture, its training using the Backpropagation algorithm, and accuracy evaluation are beyond the scope of this paper; \cite{Beohar_2021}, \cite{Krizhevsky_2012}, and \cite{Lecun_1998} provide extensive coverage of these topics. However, it is worth noting that function (\ref{triplet-loss-func:eq}) is non-smooth and non-convex, and the standard Backpropagation technique is not validated for such cases. The extension of this technique to the non-smooth non-convex case was made in \cite{Norkin_2021}.

Regarding triplet mining strategies for $(\xi_i, \xi^{+}_i, \xi^{-}_i)$, it is crucial to select an approach that produces the most informative gradients for the objective function (\ref{triplet-loss-func:eq}). Paper \cite{xuan2020} discusses various online triplet mining strategies, which select triplets within each batch of a training set on each iteration. We employed the semi-hard triplet mining strategy, which chooses an anchor-negative pair that is farther than the anchor-positive pair but within the margin $\alpha$:

\begin{equation}
    \label{semi-hard-triplet-mining:eq}
    \xi^{-}_i = \argmin_{\substack{\xi: C(\xi) \neq C(\xi_i) \\ d(f_{\theta}(\xi_i), f_{\theta}(\xi)) > d(f_{\theta}(\xi_i), f_{\theta}(\xi^{+}_i))}} d(f_{\theta}(\xi_i), f_{\theta}(\xi))
\end{equation}

\noindent where $C(\xi)$ denotes the label of an element $\xi$.

By applying ideas from \cite{Hoffer_2015,MURASAKI_ANDO_SHIMAMURA_2022,Turpault_Serizel_Vincent_2019}, we can utilize the encoded latent representations of the Triplet Network to train a Stochastic Quantization  algorithm (\ref{sgd-update-rule:eq}). This novel approach enables us to solve supervised or semi-supervised learning problems of classification on high-dimensional data. The semi-supervised learning process using the combined algorithm, assuming we have a labeled subset $\Xi^\prime \subset \Xi$ and remaining unlabeled data $\bar{\Xi}=\Xi\setminus \Xi^\prime$, can be summarized as follows:

\begin{enumerate}
    \item Train a Triplet Network $f_{\theta}$ on labeled data $\Xi^\prime$ and produce their encoded latent representations in $\mathbb{R}^n$.
    \item Utilize the trained Triplet Network $f_{\theta}$ to project the remaining unlabeled data $\bar{\Xi}$ onto the same latent representation space $\mathbb{R}^n$.
    \item Employ both labeled and unlabeled latent representations to train a Stochastic Quantization  algorithm (\ref{sgd-update-rule:eq}).
\end{enumerate}

    \section{Numerical Experiments}

To implement and train the Triplet Network, we utilized PyTorch 2.0 \cite{Ansel_2024}, a framework designed for high-performance parallel computations on accelerated hardware. The Stochastic Quantization algorithm was implemented using the high-level API of Scikit-learn \cite{Pedregosa_2011}, ensuring compatibility with other package components (e.g., model training and evaluation), while leveraging NumPy \cite{harris2020array} for efficient tensor computations on CPU. All figures presented in this study were generated using Matplotlib \cite{Hunter_2007}. The source code and experimental results are publicly available in our GitHub repository \cite{Kozyriev_2024}.

\begin{figure}
    \centering
    \includegraphics[width=\textwidth]{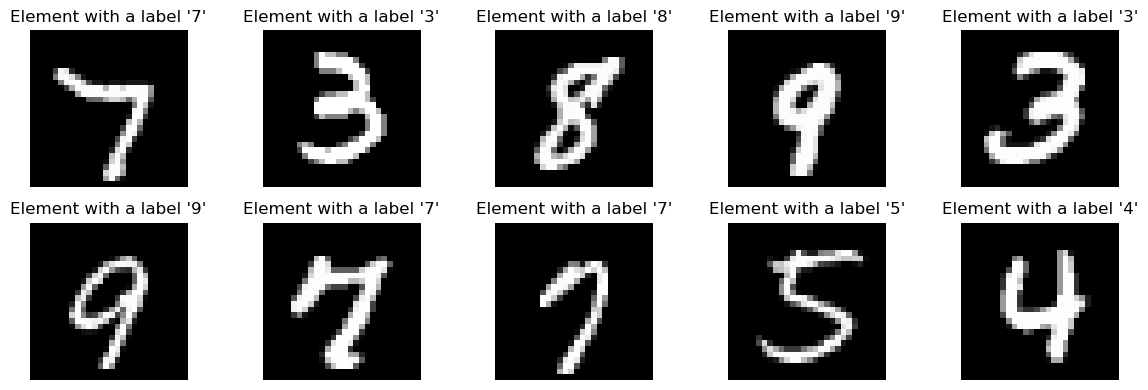}
    \caption{Representative samples from the MNIST dataset \cite{lecun2010mnist} with their corresponding labels.}
    \label{mnist:fig}
\end{figure}

For our experiments, we employed the original MNIST handwritten digit dataset \cite{lecun2010mnist} (see Fig.~\ref{mnist:fig} of representative samples from the MNIST dataset with their corresponding labels), comprising 60,000 grayscale images of handwritten digits with a resolution of 28x28 pixels, each associated with a class label from 0 to 9. Additionally, the dataset includes a corresponding set of 10,000 test images with their respective labels. It is noteworthy that we did not apply any data augmentation or preprocessing techniques to either the training or test datasets.

\begin{figure}
    \centering
    \includegraphics[width=\textwidth]{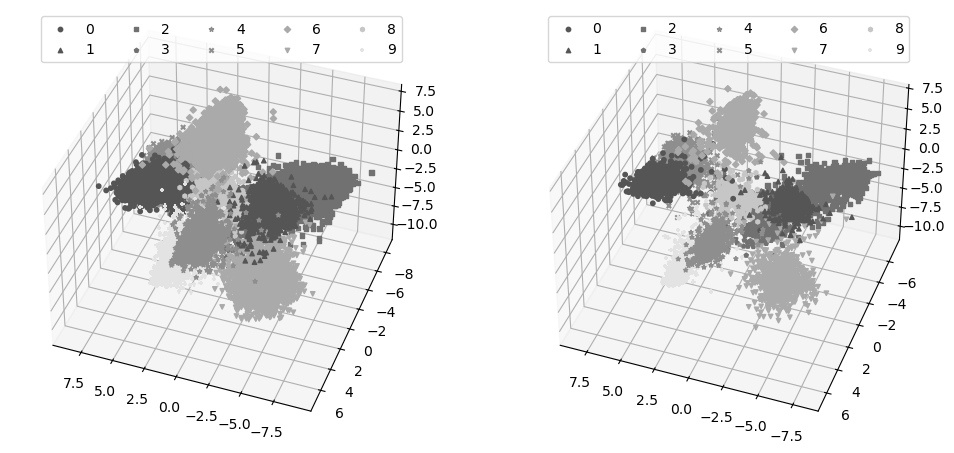}
    \caption{Latent representations of images in the train dataset (left) and test dataset (right) projected by the Triplet Network, with each element color-coded according to its label (0-9). The clustering of elements with the same label suggests that the Triplet Network successfully captured relevant features during training.}
    \label{latent-space:fig}
\end{figure}

We approached the image classification task as a semi-supervised learning problem, training models on varying fractions of labeled training data (25\%, 50\%, 75\%, and 100\%). The training dataset was split using uniform sampling into labeled and unlabeled portions according to the specified percentages. For the Triplet Network, we employed a Convolutional Neural Network architecture consisting of two Convolutional Layers with 3x3 filters (feature map dimensions of 32 and 64, respectively, with $\text{stride}=1$ and $\text{padding}=1$), followed by 2x2 Max-Pooling layers, and two Dense Layers. ReLU non-differentiable activation functions were utilized throughout the network, introducing non-linearity to each layer. Together with non-smooth Triplet loss function this makes the learning problem highly non-smooth and non-convex (see discussion of this issue in \cite{Norkin_2021}). We trained separate Triplet Network models for each labeled data fraction with the following hyperparameters: 50 epochs, batch size of 1000, learning rate $\rho = 10^{-3}$, and $l_2$ regularization rate $\lambda = 10^{-5}$. For the triplet loss (\ref{triplet-loss-func:eq}) and triplet mining (\ref{semi-hard-triplet-mining:eq}), we set the margin hyperparameter $\alpha = 1.0$. To facilitate meaningful feature capture while enabling visualization, we chose a latent space $\mathbb{R}^3$ of dimensionality $n=3$. Fig.~\ref{latent-space:fig} gives latent representations of images in the train dataset (left) and test dataset (right) projected by the Triplet Network, with each element gray-coded according to its label (0-9). The clustering of elements with the same label suggests that the Triplet Network successfully captured relevant features during training.

\begin{figure}
    \centering
    \includegraphics[width=0.8\textwidth]{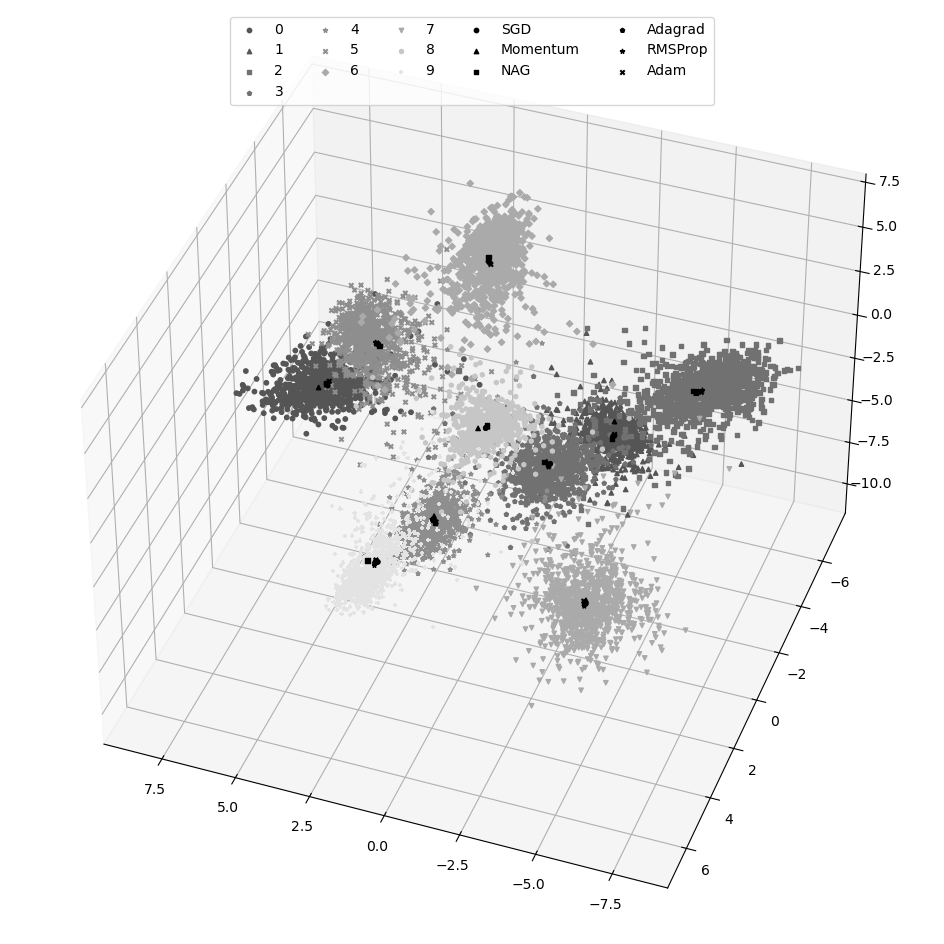}
    \caption{Optimal positions of quants for each Stochastic Quantization variant (labeled by variant name) in the latent space, relative to labeled elements (0-9) from the test dataset.}
    \label{quants:fig}
\end{figure}

\begin{figure}
    \centering
    \includegraphics[width=\textwidth]{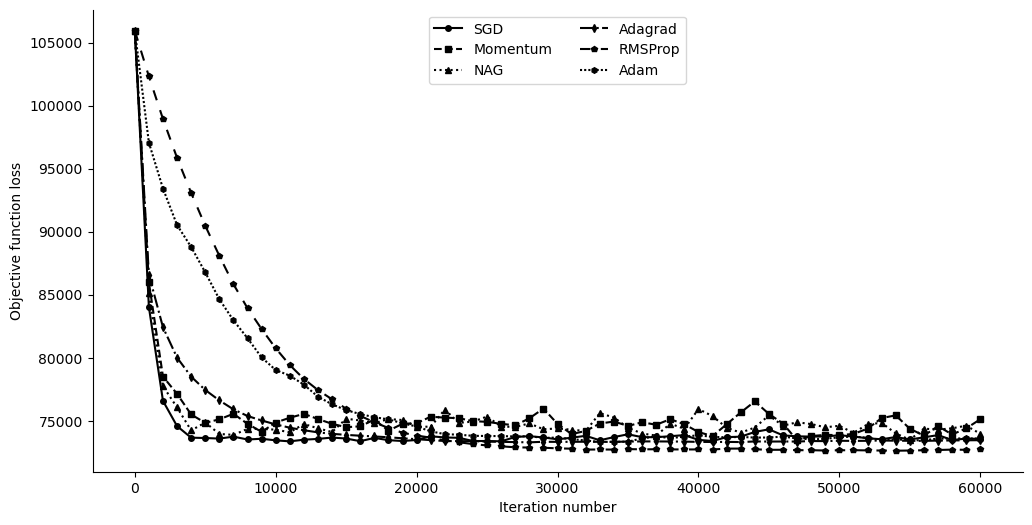}
    \caption{Convergence speed comparison of Stochastic Quantization variants on latent representations of training data with 100\% labeled fraction.}
    \label{convergence:fig}
\end{figure}

The Triplet Network was used to project latent representations onto $\mathbb{R}^3$ space from both labeled and unlabeled training data. These representations were then used to train the Stochastic Quantization algorithm as an unsupervised learning model. For each set of latent representations corresponding to different labeled data fractions, we trained the Stochastic Quantization algorithm and its adaptive variants (\ref{momentum-update-rule:eq})-(\ref{adam-update-rule:eq}) from subsection \ref{adap-stoch-quant:sec}. The initialization strategy for quant positions utilized uniform sampling from a labeled data fraction for each label. To ensure convergence, we set the rank hyperparameter to $r = 3$ and employed different learning rates for each variant: $\rho = 0.001$ for SGD, Momentum, NAG, RMSProp; $\rho = 0.1$ for AdaGrad; and $\rho = 0.01$ for ADAM. With these hyperparameters, all Stochastic Quantization variants converged to the global optima. Fig.~\ref{quants:fig} shows optimal positions of quants for each Stochastic Quantization variant (labeled by variant name) in the latent space, relative to labeled elements (0-9) from the test dataset. Fig.~\ref{convergence:fig} presents a comparative analysis of the convergence rates for the Stochastic Quantization variants by plotting the objective function value (\ref{global-sq-fn-expansion:eq}) against the iteration number $t$.

The accuracy of the trained classification models, combining Triplet Network and Stochastic Quantization, was evaluated using the F1-score metric \cite{Chinchor_1992} for weighted multi-label classification. Our experiments demonstrated that our approach achieved results comparable to state-of-the-art performance with Triplet Network and Siamese Network, as reported in \cite{Hoffer_2015}, even with limited labeled data. Table~\ref{accuracy:table} presents a comprehensive comparison of classification accuracy across different algorithms and percentages of training data.

\begin{table}
\caption{Classification accuracy comparison on MNIST dataset \cite{lecun2010mnist}.}
    \label{accuracy:table}
    \begin{tabularx}{\textwidth}{|X|*{4}{c|}}
        \hline
        \multirow{2}{*}{Algorithm} & \multicolumn{4}{c|}{Percentage of Training Data} \\
        \cline{2-5}
        & 25\% & 50\% & 75\% & 100\% \\
        \hline
        Triplet Network + KNN         &    -    &    -    &    -    & 99.54\% \\
        Siamese Network + KNN         &    -    &    -    &    -    & 97.90\% \\
        Triplet Network + SQ-SGD      & 78.10\% & 97.74\% & 97.83\% & 98.38\% \\
        Triplet Network + SQ-Momentum & 78.06\% & 97.73\% & 97.82\% & 98.36\% \\
        Triplet Network + SQ-NAG      & 78.00\% & 97.63\% & 97.72\% & 98.33\% \\
        Triplet Network + SQ-AdaGrad  & 78.95\% & 97.72\% & 97.79\% & 98.34\% \\
        Triplet Network + SQ-RMSProp  & 79.34\% & 97.70\% & 97.80\% & 98.29\% \\
        Triplet Network + SQ-ADAM     & 79.15\% & 97.70\% & 97.86\% & 98.29\% \\
        \hline
    \end{tabularx}
\end{table}

    \section{Conclusions}

In this paper, we introduced a novel approach to solving semi-supervised learning problems by combining Contrastive Learning with the Stochastic Quantization algorithm. Our robust solution addresses the challenge of scalability in large datasets for clustering problems, while also mitigating the ''curse of dimensionality'' phenomenon through the integration of the Triplet Network. Although we introduced modifications to the Stochastic Quantization algorithm by incorporating an adaptive learning rate, there is potential for further enhancement by developing an alternative modification using the finite-difference algorithm proposed in \cite{Norkin_Kozyriev_Norkin_2024}, which will be the focus of future research. The semi-supervised nature of the problems addressed in this paper arises from the necessity of using labeled data to train the Triplet Network. By employing other Deep Neural Network architectures to produce low-dimensional representations in the latent space, the semi-supervised approach of Stochastic Quantization with Encoding could be extended to unsupervised learning. Additionally, alternative objective functions for contrastive learning, such as N-pair loss \cite{Sohn_2016} or center loss \cite{qi2017}, offer further avenues for exploration in future studies. The results obtained in this paper demonstrate that the proposed approach enables researchers to train classification models on high-dimensional, partially labeled data, significantly reducing the time and labor required for data annotation without substantially compromising accuracy.

    \bibliographystyle{splncs04}
    \bibliography{references}
\end{document}